\documentclass[10pt,twocolumn,letterpaper]{article}

\usepackage{cvpr}              

\usepackage{graphicx}
\usepackage{amsmath}
\usepackage{amssymb}
\usepackage{booktabs}

\usepackage[pagebackref,breaklinks,colorlinks]{hyperref}

\usepackage[capitalize]{cleveref}
\crefname{section}{Sec.}{Secs.}
\Crefname{section}{Section}{Sections}
\Crefname{table}{Table}{Tables}
\crefname{table}{Tab.}{Tabs.}

\def\cvprPaperID{18} 
\def\confName{CVPR}
\def\confYear{2023}

\begin{document}

\title{Design a Delicious Lunchbox in Style}

\author{Yutong Zhou\\
Ritsumeikan University, Shiga, Japan\\
\href{mailto:zhou@i.ci.ritsumei.ac.jp}{\textcolor{black}{\tt\small zhou@i.ci.ritsumei.ac.jp}}
}
\maketitle

\begin{abstract}
We propose a cyclic generative adversarial network with spatial-wise and channel-wise attention modules for text-to-image synthesis~\cite{zhou2021generative}. To accurately depict and design scenes with multiple occluded objects, we design a pre-trained ordering recovery model and a generative adversarial network to predict layout and composite novel box lunch presentations~\cite{zhou2022able}. In the experiments, we devise the Bento800 dataset to evaluate the performance of the text-to-image synthesis model and the layout generation \& image composition model. 
This paper is the continuation of our previous paper works ~\cite{zhou2021generative} and ~\cite{zhou2022able}. We also present additional experiments and qualitative performance comparisons to verify the effectiveness of our proposed method. Bento800 dataset is available at \url{https://github.com/Yutong-Zhou-cv/Bento800_Dataset}.
\end{abstract}

\section{Introduction}
\label{sec:intro}

With the increasing need for portable meals as people had to leave their homes for hours, the bento box has become a popular and celebrated food culture that emphasizes flavor, visual aesthetics, nutritional balance, and convenience. Despite advances in robotics applications in food production, creating visually appealing box lunch presentations and organizing multiple food items in an orderly manner still remain a challenge for robotics. This raises a question: \textbf{What kind of composite image can provide comprehensive robotics guidance, standard placement compliance, and visually appealing presentation?}

To resolve the above issue, (1) We propose a cyclic generative adversarial network~\cite{zhou2021generative} for text-to-image generation and image captioning. (2) We introduce Bento800 ~\cite{zhou2022able}, the first manually annotated synthetic box lunch dataset. (3) We propose an aesthetic box lunch design model~\cite{zhou2022able} with pre-trained placement ordering recovery and a generative adversarial network (GAN)~\cite{goodfellow2014generative} for layout generation \& ingredients composition.
This paper effectively combines our prior works and additionally expands upon the following aspects (Details are shown in \cref{sec:details} and \cref{sec:Experiments}): (4) We rephrase text descriptions in Bento800 dataset to increase text diversity for text-to-image generation. (5) We conduct experiments to demonstrate the effectiveness of our method in synthesizing and designing novel lunchbox images.


\section{Method}
\label{sec:formatting}

\subsection{Text-to-Image Synthesis}
\label{sec:t2i}

\begin{figure}[t]
  \centering
  \includegraphics[width=0.44\textwidth,clip]{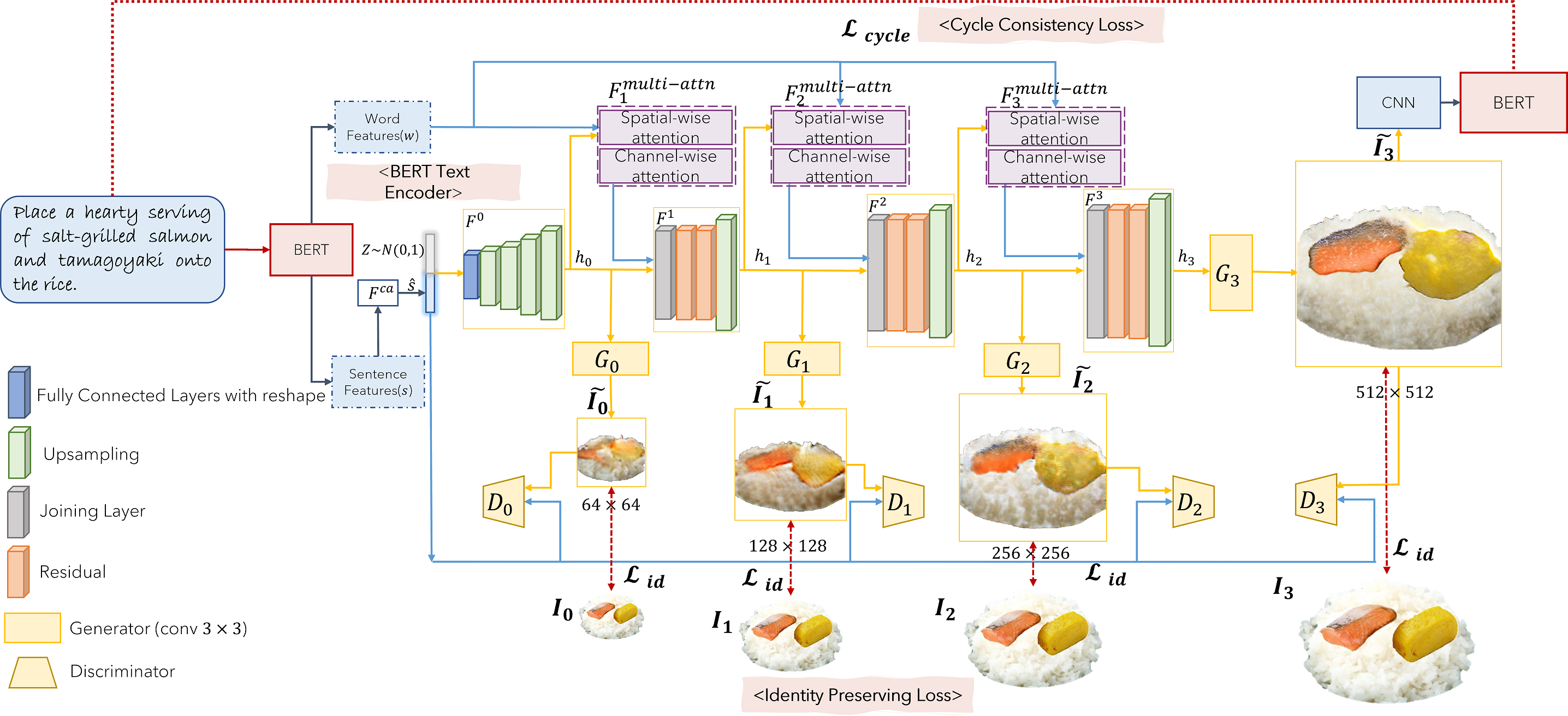}
  \caption{An overview of our proposed approach in ~\cite{zhou2021generative}.}
  \label{fig:FG}
  \vspace{-2mm}
\end{figure}

Based on our previous dual attention architecture~\cite{zhou2021}, we propose a cyclic GAN (as in \cref{fig:FG}) and adopt a pre-trained BERT base model~\cite{devlin2018bert} for text embedding. For image captioning, we employ the Inception-v3 model~\cite{Inception2016} as a CNN image encoder for extracting image features and the BERT model to generate contextualized word vectors for each caption.
The full objective function as below:
\vspace{-1mm}
\begin{equation}
  {L}_{G} = \sum_{m=1}^{M} ({L}_{G_{m}}+ {L}_{id})+ {L}_{cycle}
  \label{eq:G_all_loss}
\end{equation}
\vspace{-3mm}
\begin{equation}
  {L}_{D} = \sum_{m=1}^{M} {L}_{D_{m}}
  \label{D_all_loss}
\end{equation}
where ${L}_{G_{m}}$ and ${L}_{D_{m}}$ denote the adversarial loss at \textit{m}th stage (m=0, 1, 2, 3). The adversarial loss consists of two components: unconditional loss, which evaluates the realism of the generated image, and conditional loss, which measures the compatibility between the generated image and the input text description. ${L}_{id}$ denotes the proposed identity-preserving loss which quantifies the dissimilarity between the generated image and the ground-truth image. ${L}_{cycle}$ denotes the cycle consistency loss between predicted captions and target captions. 

\subsection{Layout Generation and Image Composition}
\label{sec:LGIC}
\begin{figure}[t]
  \centering
  \begin{minipage}[b]{\linewidth}
    \centering
    \subfloat[][Pre-processing \& Training Phase ]{\label{fig:MM1}\includegraphics[width=0.5\textwidth,clip]{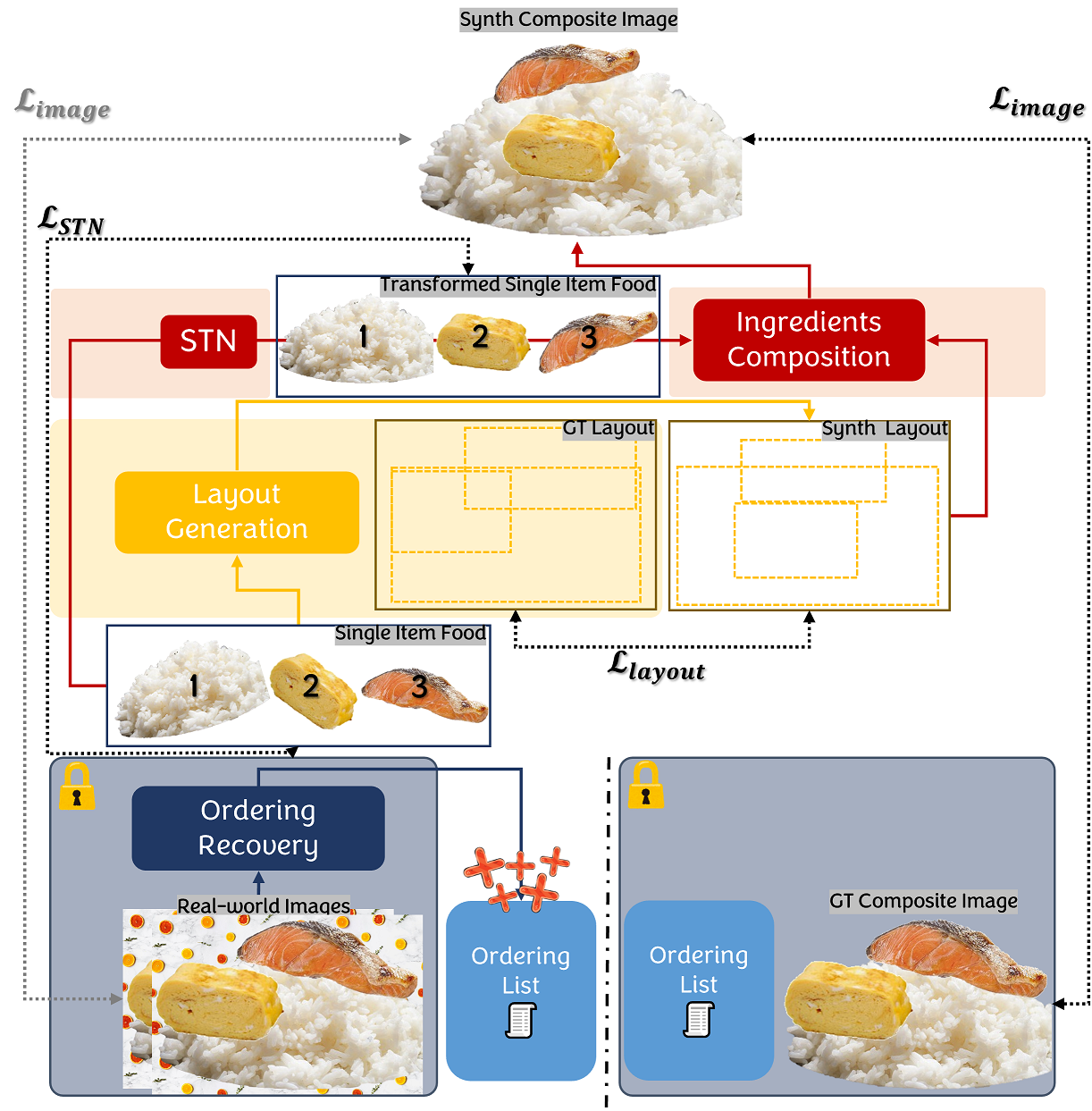}}
    \subfloat[][Testing Phase ]{\label{fig:MM2}\includegraphics[width=0.32\textwidth,clip]{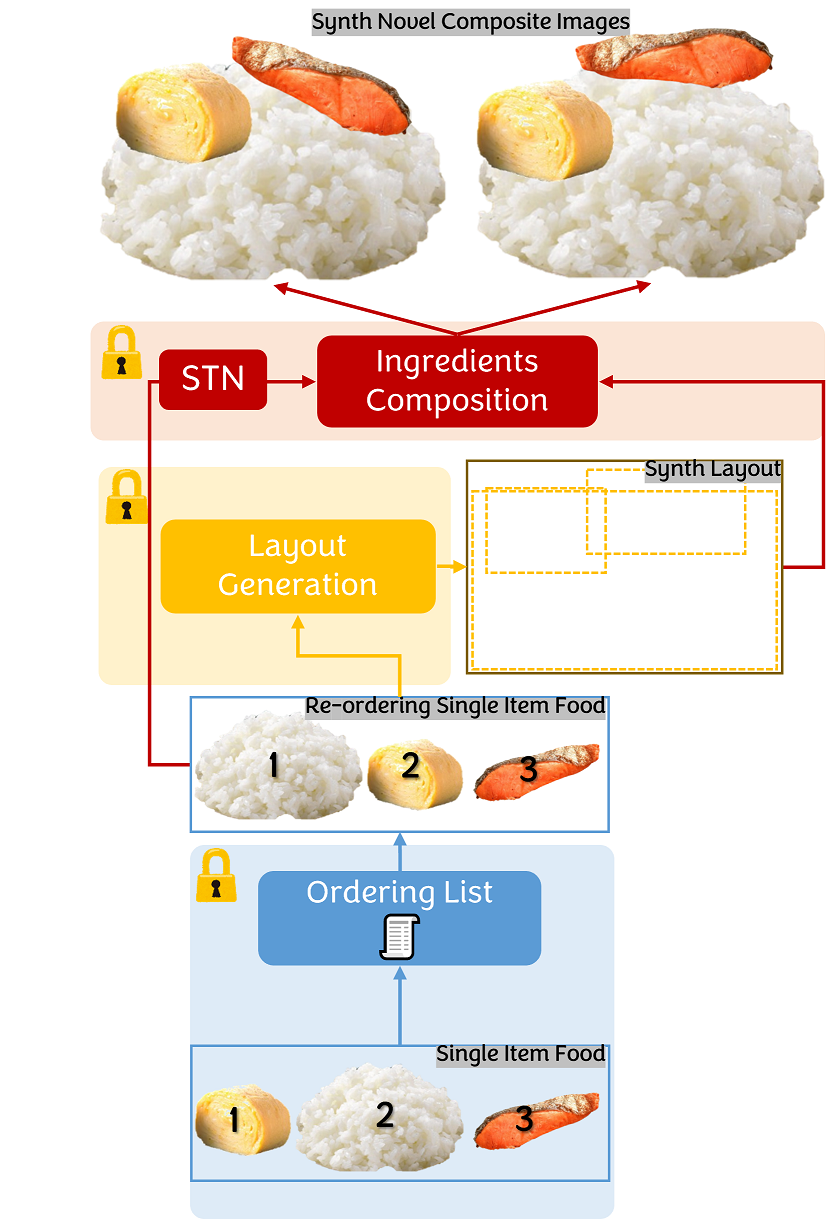}}
  \end{minipage} 
  \caption{An overview of our proposed approach in ~\cite{zhou2022able}. }
  \label{fig:MM}
  \vspace{-2.5mm}
\end{figure}

Text-guided generation models often have imprecise specifications for producing the intended target image, necessitating the use of additional controls for generating images. To achieve a precise depiction of scenes with multiple objects, we propose the following method, as shown in \cref{fig:MM}.
The pre-processing model in \cref{fig:MM1} takes synthetic or real images as input and builds a placement ordering list. During training in \cref{fig:MM1}, the layout generation model predicts layout bounding boxes while the ingredients composition model transforms food items to their corresponding transposed targets. During the testing in \cref{fig:MM2}, the weights of the layout generation model, Spatial Transformer Network (STN)~\cite{jaderberg2015spatial} model and ingredients composition model are frozen. The training objective function is computed as:
\begin{equation}
    \centering
    {L} = {L}_{layout}(G,D)+ {L}_{image}(G,D)+ {L}_{STN}
    \label{Loss}
    \vspace{-1mm}
\end{equation}
Conditional GAN loss functions (${L}_{layout}$ and ${L}_{image}$) are applied for Layout Generation and Ingredient composition network. We also adopt a differentiable grid generator and bilinear sampling to transform food items to proper positions and compute the ${L}_{STN}$ (L1 loss) between input single-item food and transformed single-item food. 

\subsection{Extend Information: Paraphrasing}
\label{sec:details}
We propose a novel dataset Bento800~\cite{zhou2022able}, which comprises 800 lunchbox images with three different types of food presentations: (1) \textit{Place fried chicken on rice}; (2) \textit{Place salt-grilled salmon and tamagoyaki on rice}; (3) \textit{Place croquette and fried shrimp on rice, fried shrimp is on croquette}. 
To produce parallel and diverse data, we use OpenAI chatGPT\footnote{\url{https://chat.openai.com/}} to rephrase the above sentences with the template “\verb+Rephrase: [Input] with 25 examples+” and “\verb+Rephrase: [Input]+"+“\verb+Please try again+"$\times24$ and also perform data cleaning. Then, we randomly select 8 sentences from the rephrased results for each lunchbox image and merge them with the original sentences to expand our dataset for text-to-image generation.

\section{Experiments}
\label{sec:Experiments}

\begin{figure}[t]
  \centering
  \includegraphics[width=0.4\textwidth,clip]{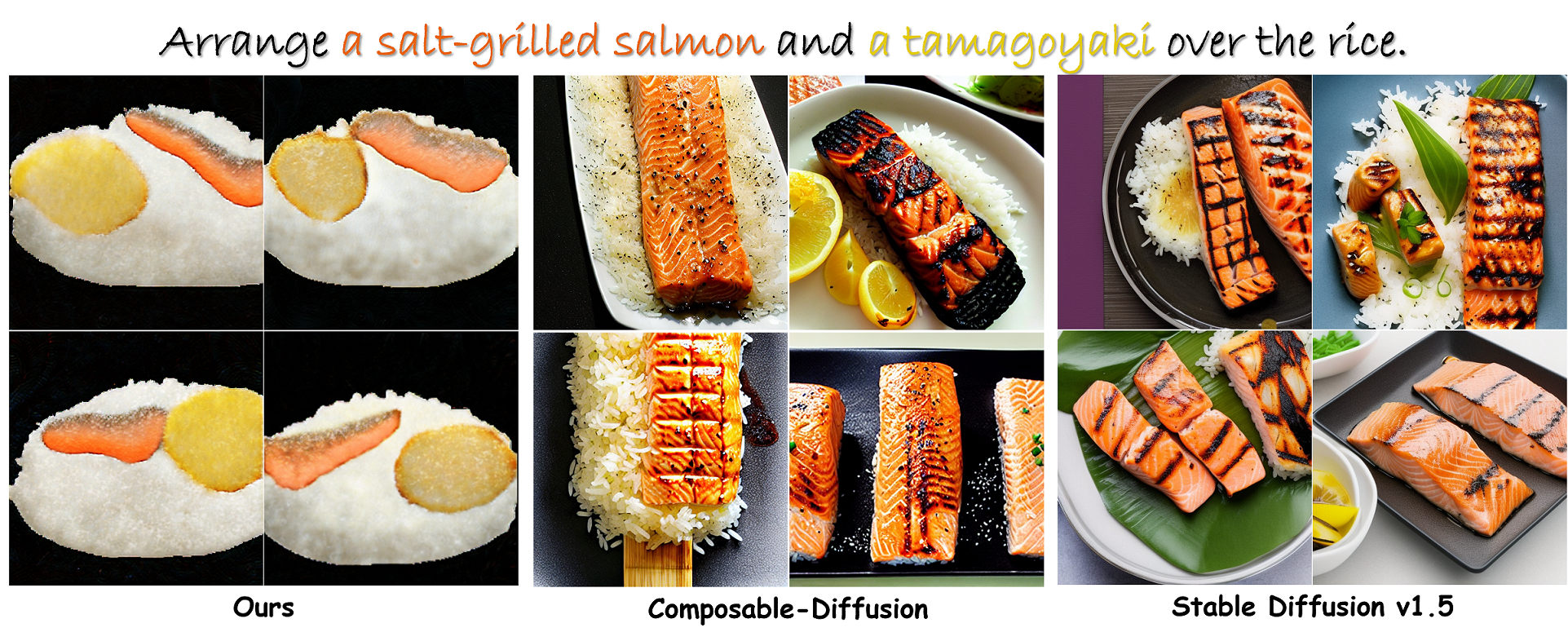}
  \vspace{-1mm}
  \caption{Qualitative comparison: Ours~\cite{zhou2021generative} (left), Composable-Diffusion~\cite{liu2022compositional} (middle) and  Stable Diffusion~\cite{rombach2022high} (right).}
  \label{fig:R_FG}
\end{figure}

\begin{figure}[t]
  \centering
  \includegraphics[width=0.4\textwidth,clip]{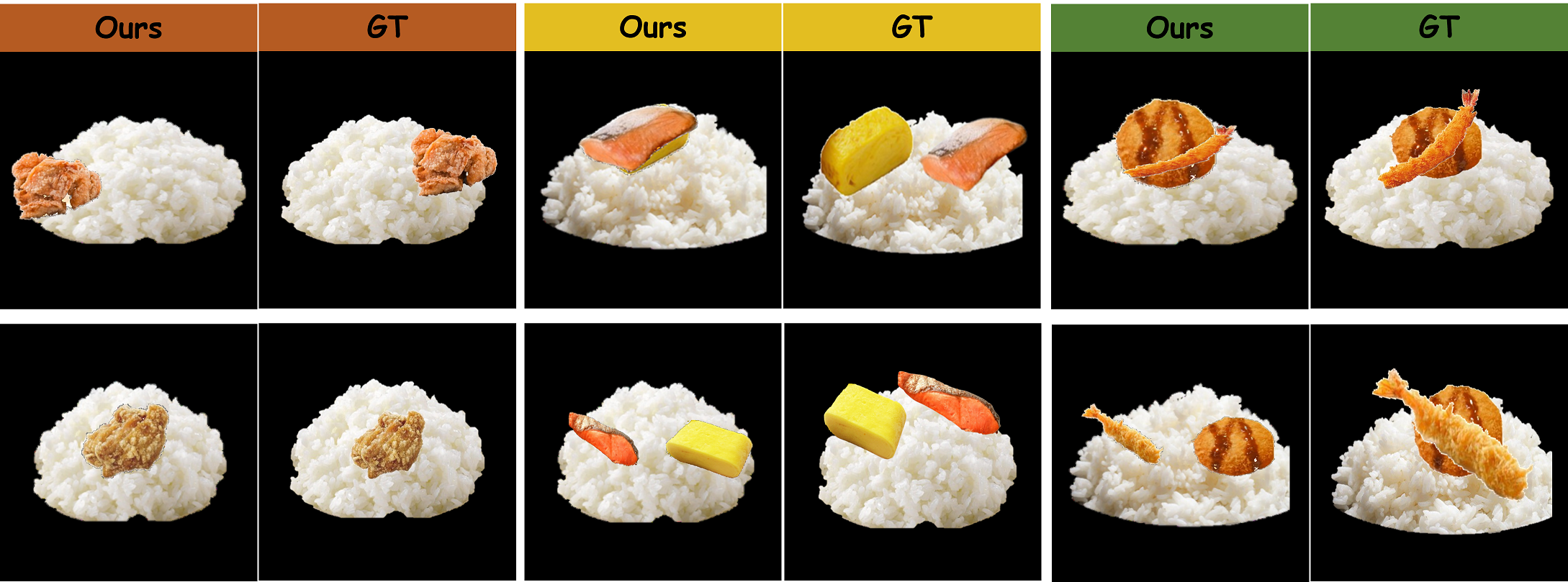}
  \vspace{-1mm}
  \caption{Our synthesized box lunch presentations~\cite{zhou2022able} (“Ours”) and human-designed ground-truth (“GT”).}
  \label{fig:R_MM1}
\end{figure}

\begin{figure}[t]
  \centering
  \includegraphics[width=0.4\textwidth,clip]{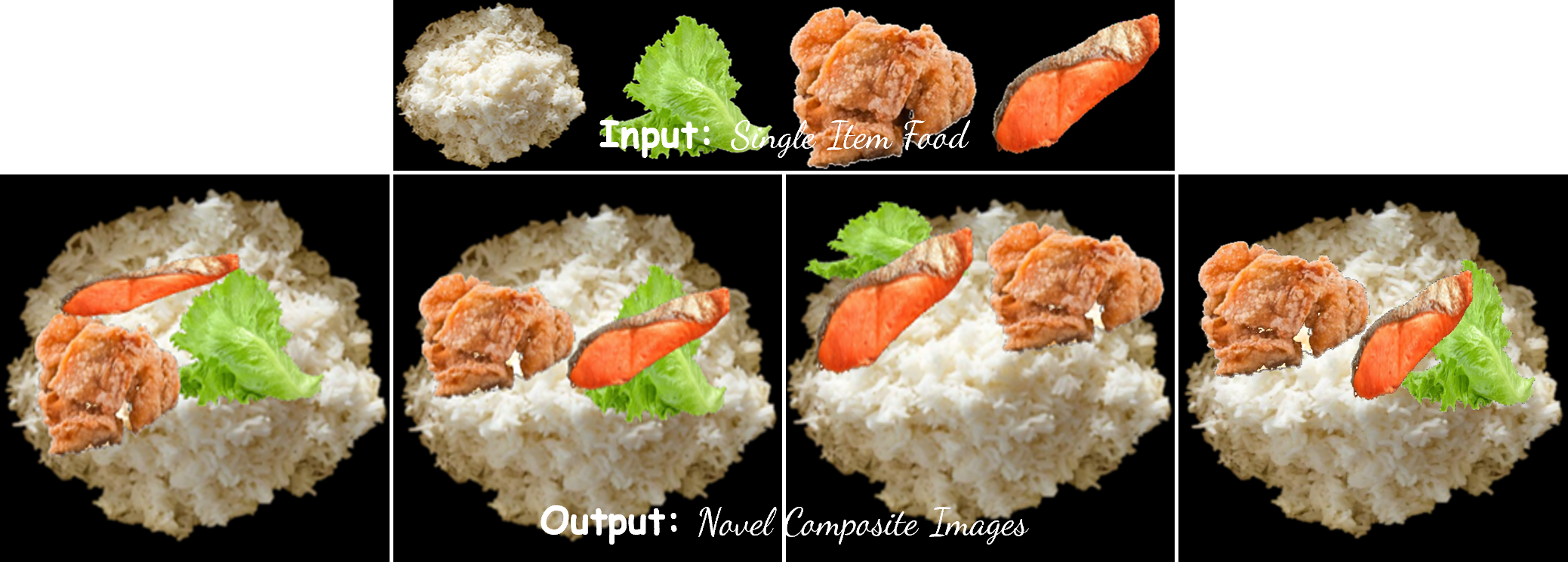}
  \vspace{-1mm}
  \caption{Diverse box lunch presentations~\cite{zhou2022able}.} 
  \label{fig:R_MM2}
  \vspace{-2.5mm}
\end{figure}

\Cref{fig:R_FG} illustrates a qualitative comparison between box lunch presentations with state-of-the-art methods for text-to-image. However, they can only synthesize poor image quality results~\cite{zhou2021generative} or images corresponding to part of the text prompt~\cite{liu2022compositional,rombach2022high}. \Cref{fig:R_MM1} demonstrates more accurate and high-quality box lunch presentations organized by our proposed method~\cite{zhou2022able} and human-designed. Three groups of results show different types of food presentations, which are also present in Bento800. \Cref{fig:R_MM2} shows various lunchboxes from single food items by randomly sampling noise. Experimental results show that our model produces diverse lunchboxes that conform to popular aesthetic preferences.

\section{Conclusion}
This paper presents a cyclic GAN for text-to-image synthesis and a layout generation \& food composition network for innovative lunchbox presentation design. We also introduce Bento800, the first manually annotated dataset with extensive annotations. For future work, we plan to create a metric that aligns with human aesthetic judgment. This exploration may encourage the potential of more varied applications for bento box presentation design in robotics.

{\small
\bibliographystyle{ieee_fullname}
\bibliography{Mybib}
}

\end{document}